\documentclass[10pt,twocolumn,letterpaper]{article}

\usepackage{cvpr}              

\usepackage{graphicx}
\usepackage{amsmath}
\usepackage{amssymb}
\usepackage{booktabs}
\usepackage{enumitem}
\usepackage{upquote}
\usepackage{listings}

\usepackage[pagebackref,breaklinks,colorlinks]{hyperref}

\usepackage[capitalize]{cleveref}
\crefname{section}{Sec.}{Secs.}
\Crefname{section}{Section}{Sections}
\Crefname{table}{Table}{Tables}
\crefname{table}{Tab.}{Tabs.}

\def\cvprPaperID{*****} 
\def\confName{CVPR}
\def\confYear{2022}

\begin{document}

\title{DeepLIIF: An Online Platform for Quantification of Clinical Pathology Slides}




\author{Parmida Ghahremani\thanks{Equal Contribution} \and Joseph Marino \footnotemark[1] \and Ricardo Dodds\footnotemark[1] \and Saad Nadeem \and \\
Memorial Sloan Kettering Cancer Center\\
{\tt\small \{ghahremp,marinoj1,doddsr,nadeems\}@mskcc.org}
}

\maketitle

\begin{abstract}
   In the clinic, resected tissue samples are stained with Hematoxylin-and-Eosin (H\&E) and/or Immunhistochemistry (IHC) stains and presented to the pathologists on glass slides or as digital scans for diagnosis and assessment of disease progression. Cell-level quantification, e.g. in IHC protein expression scoring, can be extremely inefficient and subjective. We present DeepLIIF (\url{https://deepliif.org}), a first free online platform for efficient and reproducible IHC scoring. DeepLIIF outperforms current state-of-the-art approaches (relying on manual error-prone annotations) by virtually restaining clinical IHC slides with more informative multiplex immunofluorescence staining. Our DeepLIIF cloud-native platform supports (1) more than 150 proprietary/non-proprietary input formats via the Bio-Formats standard, (2) interactive adjustment, visualization, and downloading of the IHC quantification results and the accompanying restained images, (3) consumption of an exposed workflow API programmatically or through interactive plugins for open source whole slide image viewers such as QuPath/ImageJ, and (4) auto scaling to efficiently scale GPU resources based on user demand.    
\end{abstract}

\section{Introduction}
\label{sec:intro}
\begin{figure}[t!]
\centering
\setlength{\tabcolsep}{1pt}
\begin{tabular}{ccc}
\includegraphics[width=0.16\textwidth]{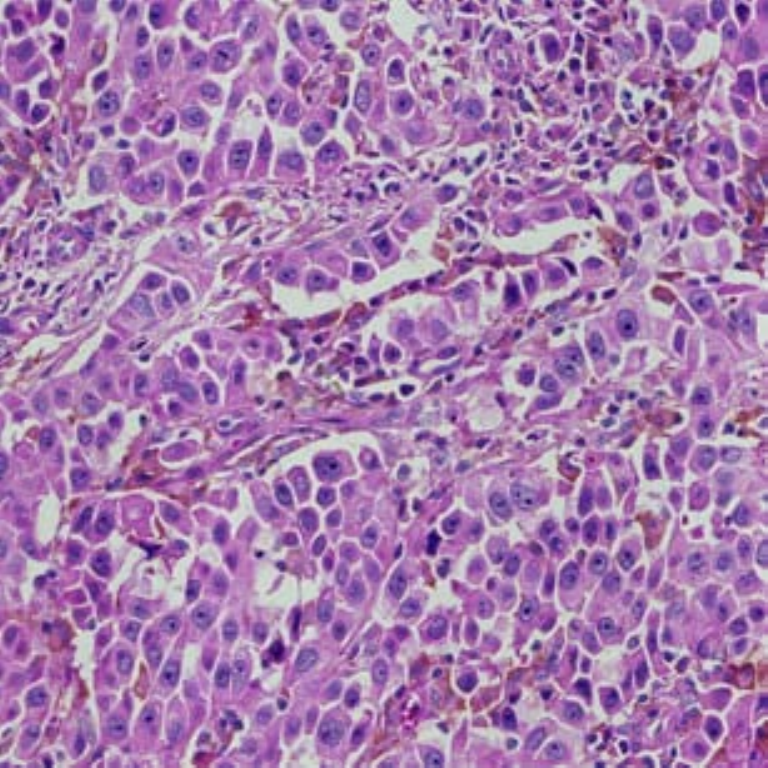}&
\includegraphics[width=0.16\textwidth]{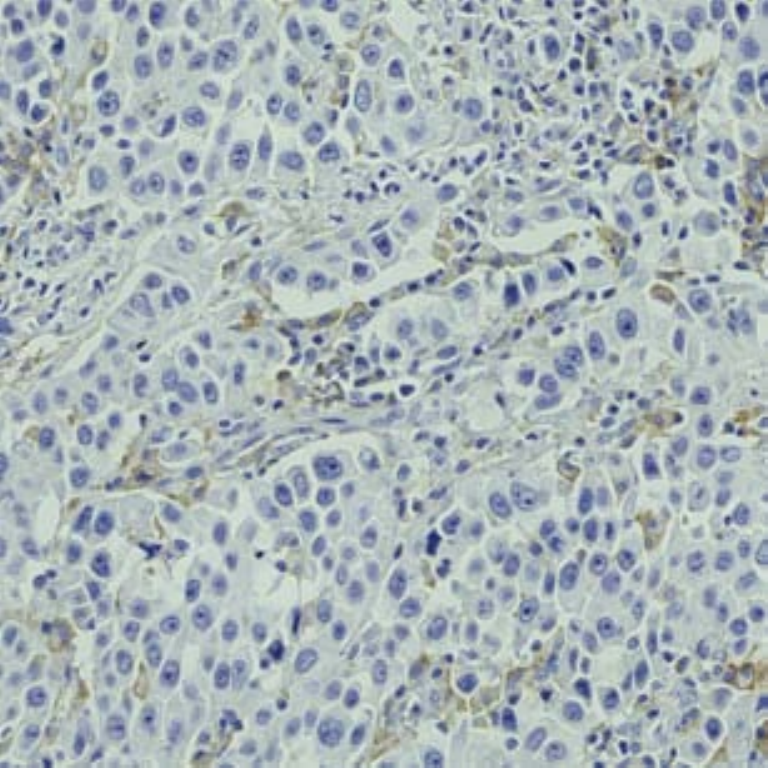}&
\includegraphics[width=0.16\textwidth]{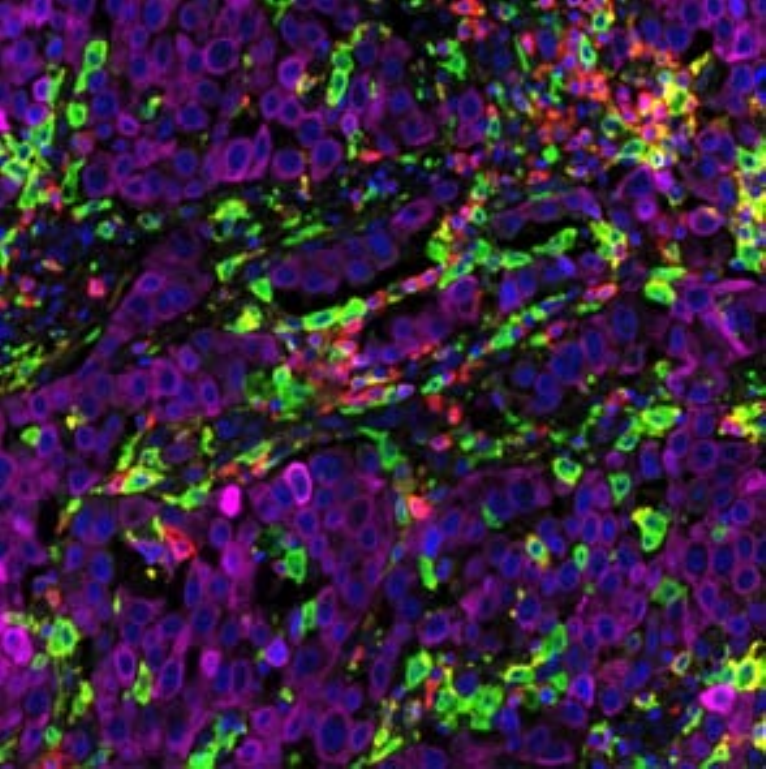}\\
H\&E (clinic) & IHC (clinic) & Multiplex (R\&D)
\end{tabular}
\caption{Brightfield Hematoxylin-Eosin (H\&E) and singleplex/protein immunohistochemistry (IHC) stains are cheap, high throughput, and clinically deployed. Multiplex IHC staining, still in the research phase, allows detection and co-visualization of multiple protein markers, providing deeper characterization of tissue microenvironment. DeepLIIF bridges the gap between clinic and research to create more informative representations for diagnosis and assessment of disease progression.}
\label{fig:pathology_slides}
\end{figure}

\noindent
Single cell protein expression quantification by pathologists and researchers in digital pathology serves a critical role in characterizing tissue microenvironment (TME) for clinical diagnostics to guide patient therapy as well as for developing research predictive and prognostic biomarkers. However, there are widely reported variations in manual singleplex/protein immunohistochemistry (IHC) scoring, even between experienced pathologists, that can negatively impact patient outcomes. Research multiplex imaging platforms can overcome these variations by allowing detection and co-visualization of multiple protein markers in a single tissue sample for more precise scoring of protein markers-of-interest in TME.  An example of the visual difference between H\&E, IHC, and multiplex stains is shown in Figure~\ref{fig:pathology_slides}.

\begin{figure*}[t]
\centering
\includegraphics[width=0.98\textwidth,keepaspectratio]{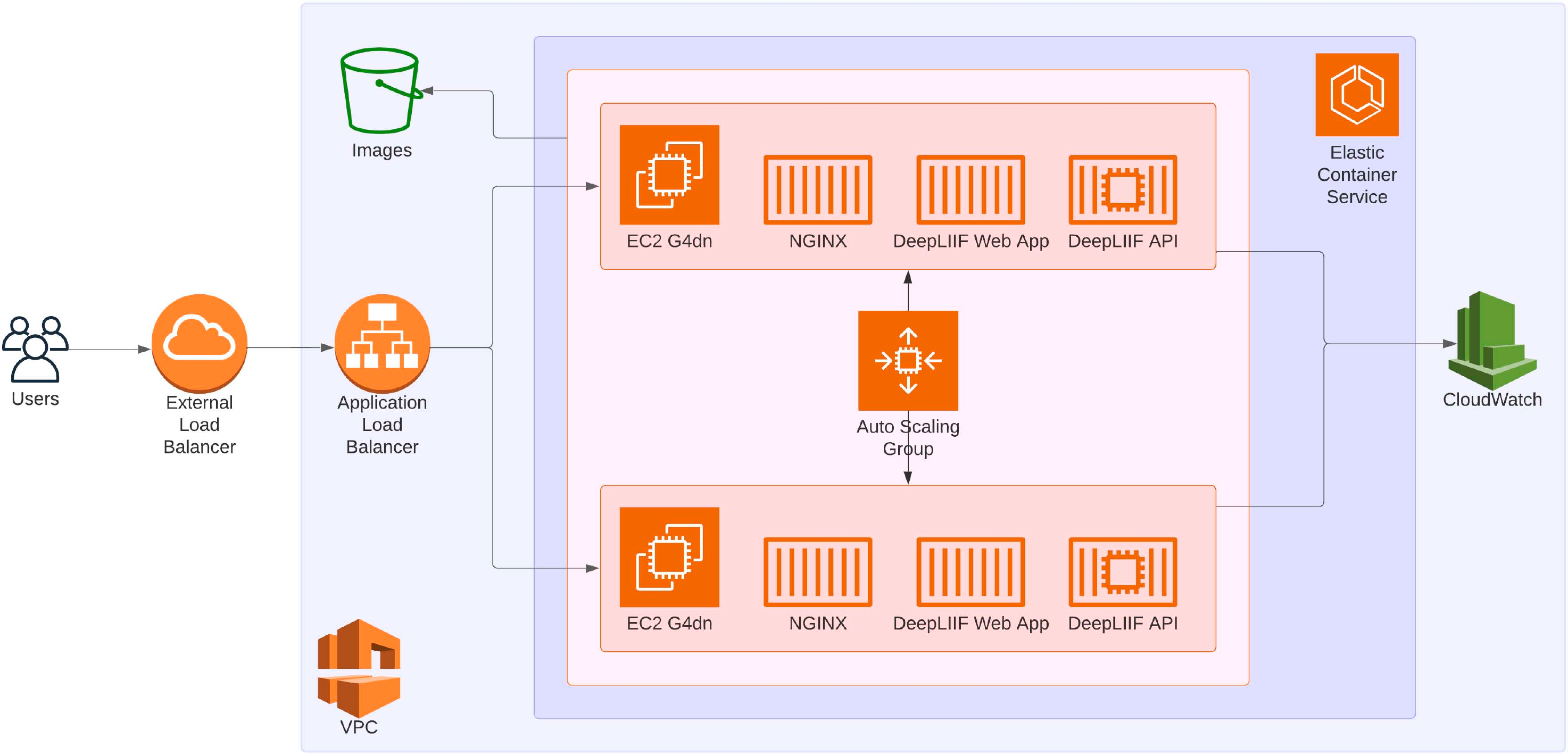}
\caption{System design diagram for the DeepLIIF online platform deployed on AWS.  The application is running on an Elastic Container Service (ECS) cluster with auto scaling.  The Web App service handles requests through the https://deepliif.org website, and uses an S3 bucket to store images uploaded by the user, along with the resultant inference images and post-processed images.  The API service runs the actual DeepLIIF model and performs post-processing on the images.  The Web App communicates with the API service running on the same machine.  Direct programmatic access to the API is possible, allowing for the creation of an ImageJ plugin.}
\label{fig:system}
\end{figure*}

There are several open source tools for segmenting cells in pathology slides stained with H\&E (Hematoxylin \& Eosin), IHC (Hematoxylin + brown DAB substrate), and multiplex \cite{lucas2021open}. These tools are available as (1) web apps (easiest to run with no prerequisite computational expertise), (2) plugins for image analysis toolboxes (mandates familiarity with these toolboxes), (3) coding notebooks (requires basic coding skills), and (4) code-based pipelines (requires significant coding expertise). In this demonstration paper, we focus on tools that are available as web apps and hence are accessible to a broad audience, including both image analysis experts and non-experts. 

There are currently three tools that are available as web apps, namely DeepCell (\url{https://deepcell.org/}) \cite{moen2019deep,greenwald2021whole}, Cellpose (\url{https://www.cellpose.org/}) \cite{stringer2021cellpose}, and NucleAIzer (\url{https://www.nucleaizer.org/}) \cite{hollandi2020nucleaizer}. DeepCell and Cellpose focus on nuclei and cytoplasm segmentation in multiplex images whereas NucleAlzer can segment nuclei across H\&E, IHC, and multiplex images (via style transfer). DeepCell and Cellpose can also perform H\&E and IHC nuclei segmentation if these are provided as grayscale images. DeepCell is the only tool among the above three that allows interaction with the segmented results via the DeepCell-Label web module (\url{https://label.deepcell.org/}). The Cellpose web app allows $512\times512$ input patches whereas DeepCell restricts input to $2048\times2048$ pixels in order to output results in a reasonable amount of time (less than a minute).  

None of the above web apps perform both cell segmentation and classification simultaneously. In brightfield H\&E and IHC slides, the two channels (Hematoxylin and Eosin in H\&E and Hematoxylin and DAB in IHC) cannot be visualized or analyzed separately without performing the necessary digital stain deconvolution pre-processing step. This is not the case with the research multiplex immunofluorescence platforms, for example, that output each marker as an independent channel that can be visualized and analyzed separately or as composites. Leveraging this insight, we developed DeepLIIF (published in \textit{Nature Machine Intelligence} \cite{ghahremani2021deepliif}) for virtual multiplex immunofluorescence restaining of standard IHC slides that performs stain deconvolution and cell segmentation/classification in a single step to output clinically relevant IHC scores (mostly quantified as relevant DAB brown cells divided by the total cells). We showed that DeepLIIF, trained on co-registered IHC and multiplex immunofluorescence images, not only achieves state-of-the-art results in IHC nuclear protein marker scoring (Ki67, ER, PR, P53) but also works out-of-the-box for H\&E nuclei segmentation as well as cytoplasmic markers (that are expressed close to the nuclei, e.g. CD3/CD8).

This demonstration paper accompanies our \textit{Nature Machine Intelligence} manuscript \cite{ghahremani2021deepliif}. Specifically, here we present the DeepLIIF cloud-native platform which supports (1) more than 150 proprietary/non-proprietary input formats via the Bio-Formats standard, (2) interactive adjustment, visualization, and downloading of the IHC quantification results and the accompanying restained images, (3) consumption of an exposed workflow API programmatically or through interactive plugins for open source whole slide image viewers such as QuPath/ImageJ, and (4) auto scaling to automatically and efficiently scale GPU resources based on user demand.

\section{System Design}
\label{sec:design}

\begin{figure*}[t]
\centering
\begin{subfigure}{.33\textwidth}
\raggedright
\frame{\includegraphics[width=.995\linewidth,keepaspectratio]{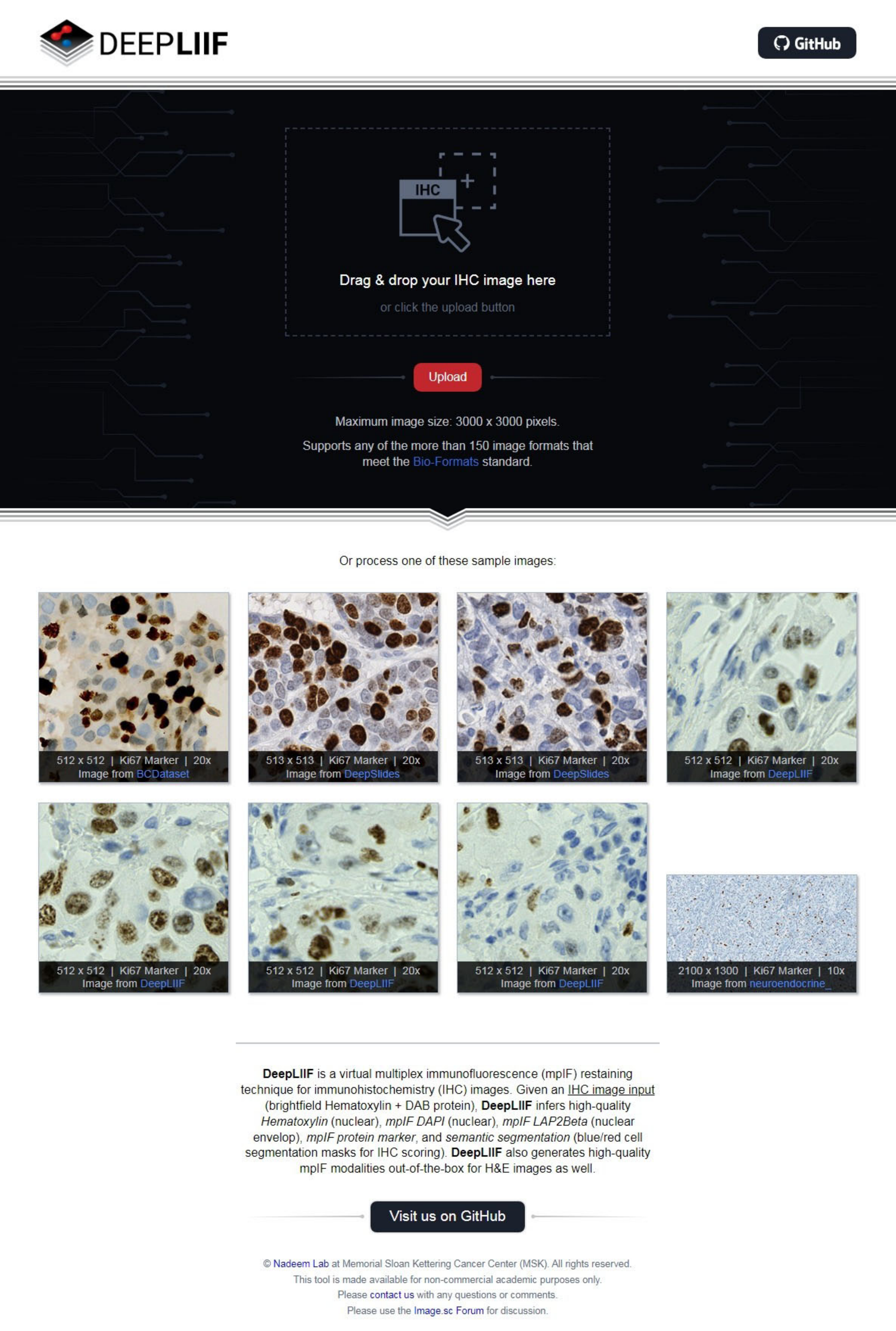}}
\caption{Image upload.}
\label{fig:website-upload}
\end{subfigure}
\begin{subfigure}{.33\textwidth}
\centering
\frame{\includegraphics[width=.995\linewidth,keepaspectratio]{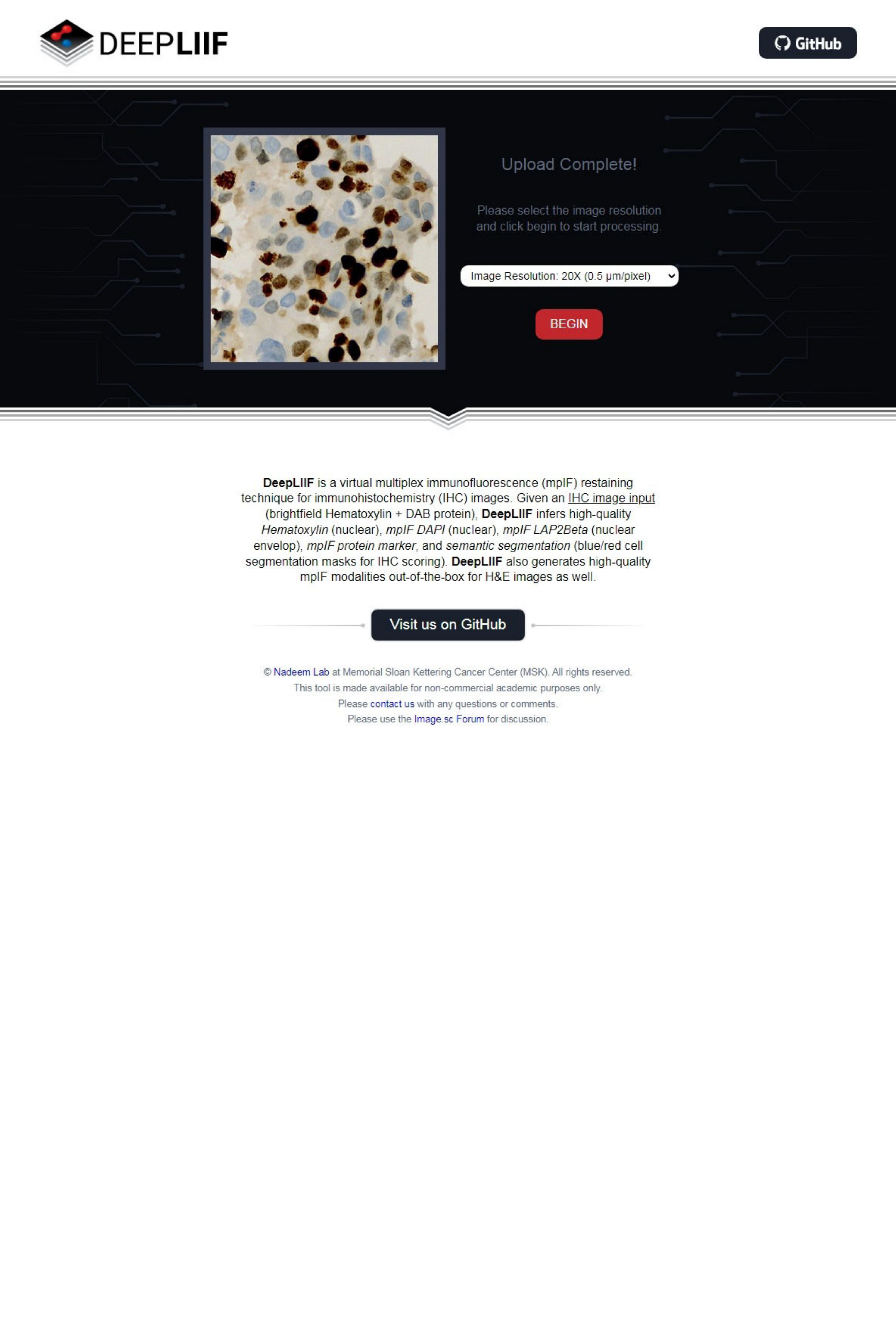}}
\caption{Image verification.}
\label{fig:website-verify}
\end{subfigure}
\begin{subfigure}{.33\textwidth}
\raggedleft
\frame{\includegraphics[width=.995\linewidth,keepaspectratio]{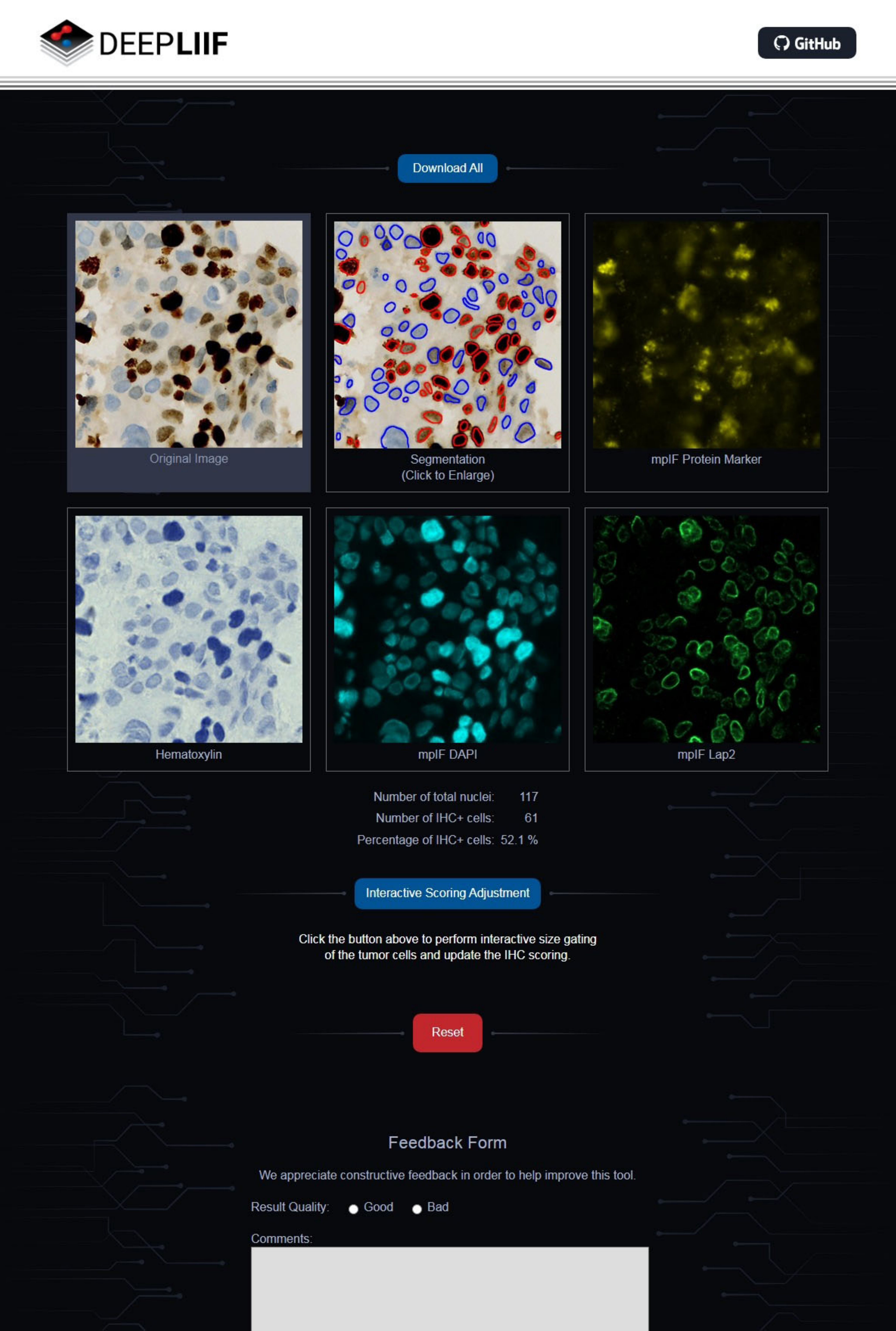}}
\caption{Results view.}
\label{fig:website-results}
\end{subfigure}
\caption{DeepLIIF website interface. (a) The DeepLIIF homepage, where the user has the option to upload their own image, or select one of several sample images to process.  (b) Choice of resolution ($10\times$, $20\times$, or $40\times$) along with preview of uploaded image.  (c) Results page, showing the inferred modalities, segmentation, and IHC quantification score.  The user can also download the full resolution inferred images and quantification scores, as well as leave feedback.}
\label{fig:website}
\end{figure*}

Our DeepLIIF cloud native platform is hosted on Amazon Web Services (AWS).  It is deployed using an Elastic Container Service (ECS) cluster powered by an Auto Scaling group to automatically adjust resources to demand.  An Amazon S3 bucket is used for cloud storage for images uploaded through the website interface and the corresponding results.  An overview of our system design architecture is shown in Figure~\ref{fig:system}.

We have three services, each running in Docker Containers for ease of portability:
\begin{itemize}
    \item Web App service for front-end website content.
    \item API service for back-end data processing.
    \item Nginx reverse proxy to serve requests.
\end{itemize}

Both the web application and the API are packaged using Flask.  The API service requires use of a CUDA-enabled NVIDIA GPU with at least 4GB of video memory in order to run the DeepLIIF inference model.  To meet this requirement, the Auto Scaling group spins up Elastic Compute Cloud (EC2) G4dn machines, each with a single GPU.  Each machine that is started contains all three services, with the web service communicating with the API service on the same machine.

\paragraph{Web App Service.}
The Web App service serves all of the front end website content for each page of our web application.  Additionally, user input is handled by this module, including verifying that uploaded files are valid images and within the specified limit of $3000\times3000$ pixels.  All connections to our S3 cloud storage bucket occur from this service, including both the original uploaded image and the inference results returned by the API service.

\paragraph{API Service.}
The API service handles all of the back end computational tasks, including the DeepLIIF model, segmentation post-processing, and quantification.  Splitting these computational tasks into a separate service allows the website interface to remain responsive to user requests even during long running processes.  In addition to providing back-end service for the DeepLIIF website, the API can also be accessed programmatically, allowing users to write and integrate their own code.  This capability has also facilitated our ability to develop a plugin that can be used to access DeepLIIF directly from within the ImageJ application.

\paragraph{Input Images.}
We utilize Bio-Formats~\cite{linkert2010bioformats} for reading the input image, which allows our application to support over 150 image formats, including all standard formats in the digital pathology domain.  To maintain an acceptable processing time and ensure equitable usage of resources, input data is currently limited to images of size no greater than $3000\times3000$ pixels.  Since the website handles a series of separate requests across a session, the image data is stored in an S3 bucket.  Similarly, the result images and quantification are stored in the S3 bucket to allow for interaction and downloading of the complete result package.  Images uploaded to the API are not stored in the cloud (and neither are the corresponding results), since the API consumer is responsible for handling all of the data.

\paragraph{License Agreement}
Upon the first visit to the DeepLIIF website in a session, the user is presented with the Terms of Use for the website and results.  Memorial Sloan Kettering Cancer Center (MSK) makes no warranties as to the accuracy of the results.  Submitted images must be the property of the user uploading said images, and MSK reserves the right to copy and use any submitted images as desired.  Submitted images should not contain and personally identifiable information (PII) or personal health information (PHI).  DeepLIIF content and results are for personal and academic research only, and may not be used in any commercial setting.  The underlying DeepLIIF project (\url{https://github.com/nadeemlab/DeepLIIF}) is licensed under the Apache 2.0 with Commons Clause license, and is available for non-commercial academic purposes.

\section{User Experience}
\label{sec:uesr}
\noindent
We provide three methods with which users can access the capabilities of the online DeepLIIF platform.  The first is through an ordinary website interface, where users can upload images, view/download results, and visualize and interact to adjust the quantification score.  For users seeking a more automated approach, we have exposed an API endpoint which can be accessed programmatically.  Additionally, we have utilized this endpoint to create an ImageJ plugin, allowing users to access DeepLIIF directly within the ImageJ application.

\subsection*{Website Interface}
\noindent
The DeepLIIF platform is made available through a conventional website user interface, which requires only a standards-compliant modern web browser and no other client software to be installed.  The steps in using the website to upload and obtain the results from the DeepLIIF model are shown in Figure~\ref{fig:website}.  These figures show the appearance of the site for desktop users, though the interface is fully responsive and can be used on any device.  The general steps for a user on the website include:
\begin{enumerate}
    \item Upload a digital pathology image.
    \item Verify thumbnail and select image resolution.
    \item View/download image and quantification results.
    \item Interactively adjust the segmentation and update the final results.
\end{enumerate}

\paragraph{Image Upload.}
The homepage for DeepLIIF (Figure~\ref{fig:website-upload}) provides the user with the option to upload an image either through drag and drop on the target area or via a File Upload dialog box.  Users can also choose from one of the sample images below the upload area if they want to try out the system but do not have their own pathology images.  Pre-calculated results are not used for the sample images, so if a sample is selected, it is processed through the same pipeline as any image uploaded by a user would be.  Once a file or sample is selected, the image is automatically sent to the server for verification.

\paragraph{Image Verification.}
Once the uploaded image has been verified by the server as an image and within the required size (currently limited to $3000\times3000$ pixels), the image data (without any metadata from the original upload) is stored in our S3 bucket.  A thumbnail is generated and returned for display in the interface (Figure~\ref{fig:website-verify}).  The user can verify that the image appears as expected, and is given the option to choose the resolution (magnification level) used when the image was captured ($10\times$, $20\times$, or $40\times$).  Once the user is ready, processing through DeepLIIF begins.

\paragraph{Results View.}
Once the DeepLIIF model has run and post-processing is completed, the results view is presented to the user (Figure~\ref{fig:website-results}).  Thumbnails of the result inferred modalities (and original upload) are displayed, along with the classified segmentation outlines overlaid on the original image.  The inferred modalities that are displayed include the protein marker, hematoxylin, DAPI, and Lap2 images.  Beneath these thumbnails are the scoring results, including the total number of nuclei, the number of positive cells, and the percentage of positive cells.  For further inspection and records, all of the full resolution images (along with the scoring results) can be downloaded in a single ZIP file.  At the bottom of the page, a feedback form is provided so that the user can leave comments specifically about the image that was processed, which will allow us to retrospectively evaluate the performance of our system.

\begin{figure*}[t]
\centering
\begin{subfigure}{.33\textwidth}
\raggedright
\frame{\includegraphics[width=.995\linewidth,keepaspectratio]{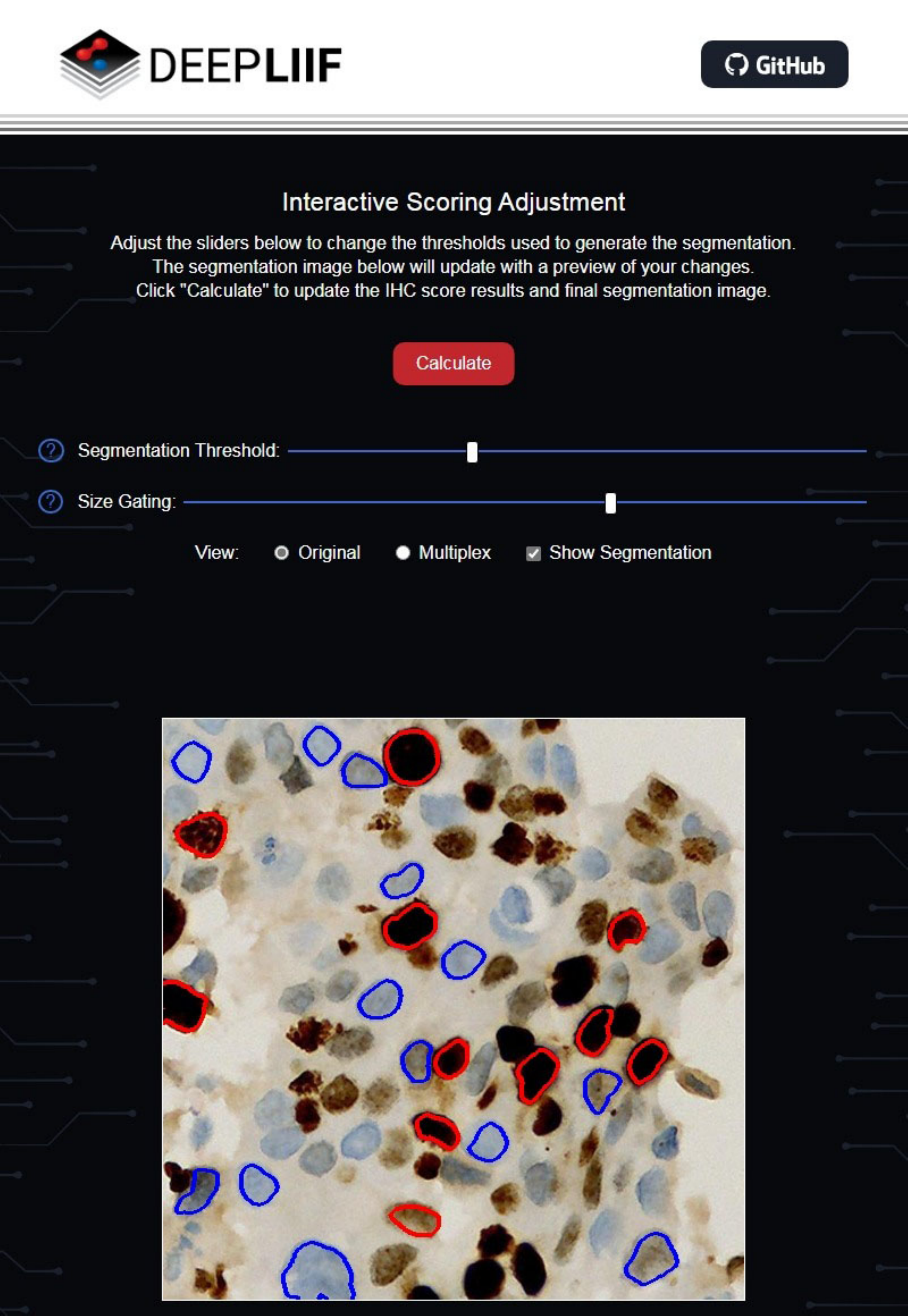}}
\caption{Original IHC with segmentation.}
\label{fig:interaction-original}
\end{subfigure}
\begin{subfigure}{.33\textwidth}
\centering
\frame{\includegraphics[width=.995\linewidth,keepaspectratio]{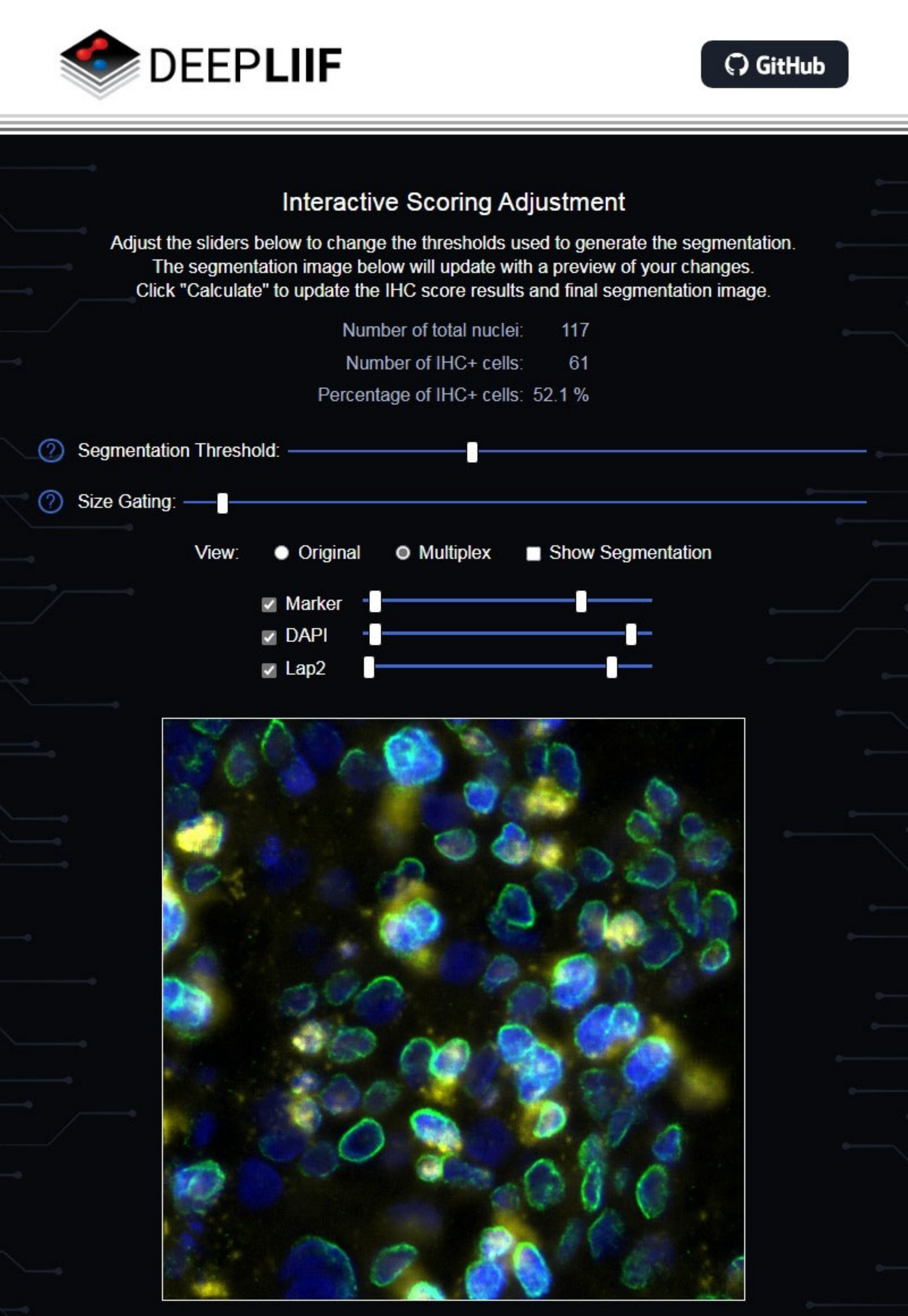}}
\caption{Inferred multiplex DAPI + Ki67 + Lap2.}
\label{fig:interaction-dapi-ki67-lap2}
\end{subfigure}
\begin{subfigure}{.33\textwidth}
\raggedleft
\frame{\includegraphics[width=.995\linewidth,keepaspectratio]{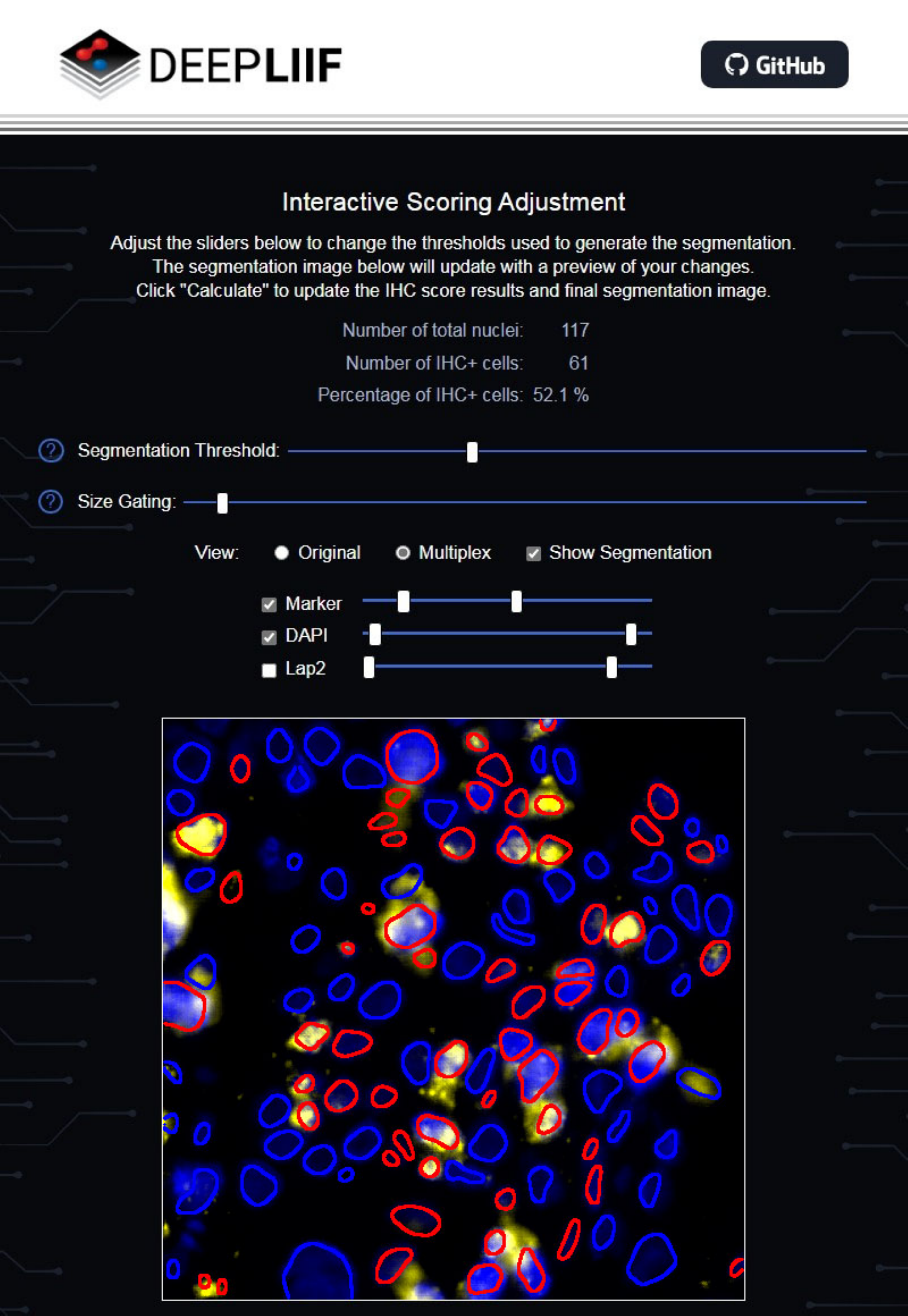}}
\caption{Inferred DAPI + Ki67 with segmentation.}
\label{fig:interaction-dapi-ki67}
\end{subfigure}
\caption{User interaction and multiplex visualization.  The user can interactively adjust the segmentation threshold and perform size gating.  The on-screen preview will update in real time to show the new cell segmentation based on the adjustments, and the IHC score can then be updated with these user-specified parameters.  User controlled visualization of the inferred multiplex data is also provided.  (a)~Original IHC image with segmentation outlines.  (b) Inferred multiplex image of DAPI + Ki67 + Lap2.  (c) Inferred multiplex image of DAPI + Ki67 with segmentation outlines.}
\label{fig:interaction}
\end{figure*}

\paragraph{Interactive Adjustment.}
In addition to the automatic results provided by DeepLIIF, our online platform provides an interactive tool with which the user can adjust the segmentation results, and thus the quantification results.  This view also allows the user to visualize the results in multiple ways, such as with or without segmentation outlines and various combinations of inferred multiplex data.  The user can adjust the segmentation threshold and perform size gating on the image.  A visual preview of the adjustments is shown in the page as the sliders are moved, implemented in JavaScript and using web workers for distributed processing to yield real-time performance even at the highest supported image resolution of $3000\times3000$.  When the user is satisfied with the preview segmentation overlay, a button is clicked to perform the full update on the server for both the segmentation image and the quantification.  The user can also download these updated results, if desired.  The interface for this interactive adjustment is shown in Figure~\ref{fig:interaction}.

\paragraph{Multiplex Visualization}
In addition to the segmentation overlay on the original IHC image, we also provide an inferred multiplex view which is user controlled.  The user can combine and adjust the following three inferred multiplex modalities, which are part of the DeepLIIF results:
\begin{itemize}
    \item Protein marker, indicating IHC positive cells.
    \item DAPI stain, highlighting cell locations.
    \item Lap2 stain, highlighting cell boundaries.
\end{itemize}

These inferred multiplex images help the user to view the cellular data without background noise and evaluate the accuracy of the segmentation results.  Each of these modalities can be toggled on or off, and the double slider can be used to adjust the histogram normalization window for each marker/stain channel. Additionally, the user can toggle the segmentation outlines on or off for both the inferred multiplex view and the original uploaded image view.

As an example, the combination of DAPI + Lap2 + Ki67 without the segmentation outlines (Figure~\ref{fig:interaction-dapi-ki67-lap2}) allows the user to compare against the original IHC image to confirm that the DeepLIIF model is accurately identifying the individual cells along with protein markers. The combination of DAPI + Ki67 with the segmentation outlines (Figure~\ref{fig:interaction-dapi-ki67}) allows the user to confirm the accuracy of the cell segmentation and classification.

\begin{figure*}[t]
\centering
\begin{subfigure}{.28214774\textwidth}
\raggedright
\includegraphics[width=.99\linewidth,keepaspectratio]{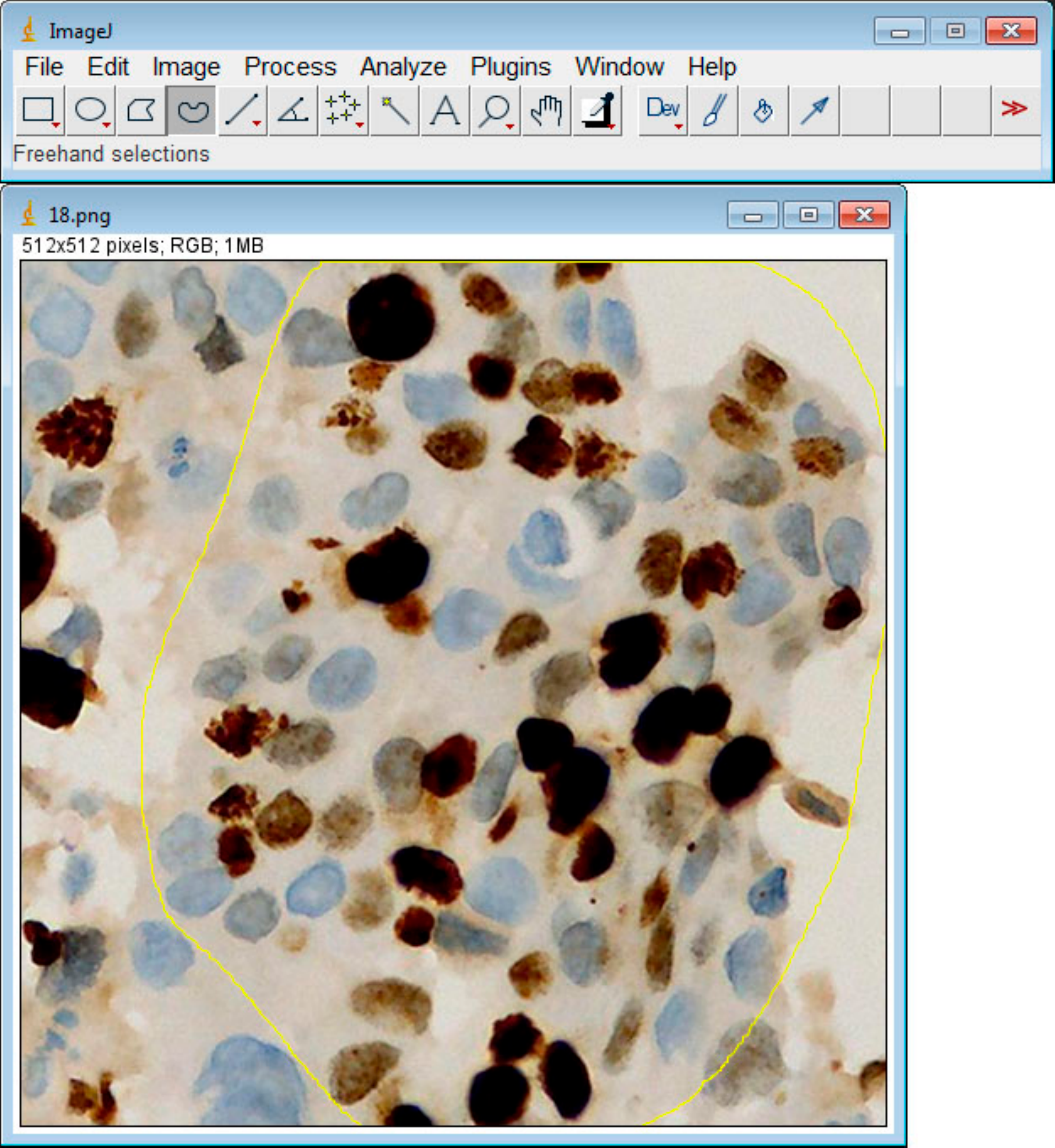}
\caption{Select a region of interest (ROI).}
\end{subfigure}
\begin{subfigure}{.33451436\textwidth}
\centering
\includegraphics[width=.99\linewidth,keepaspectratio]{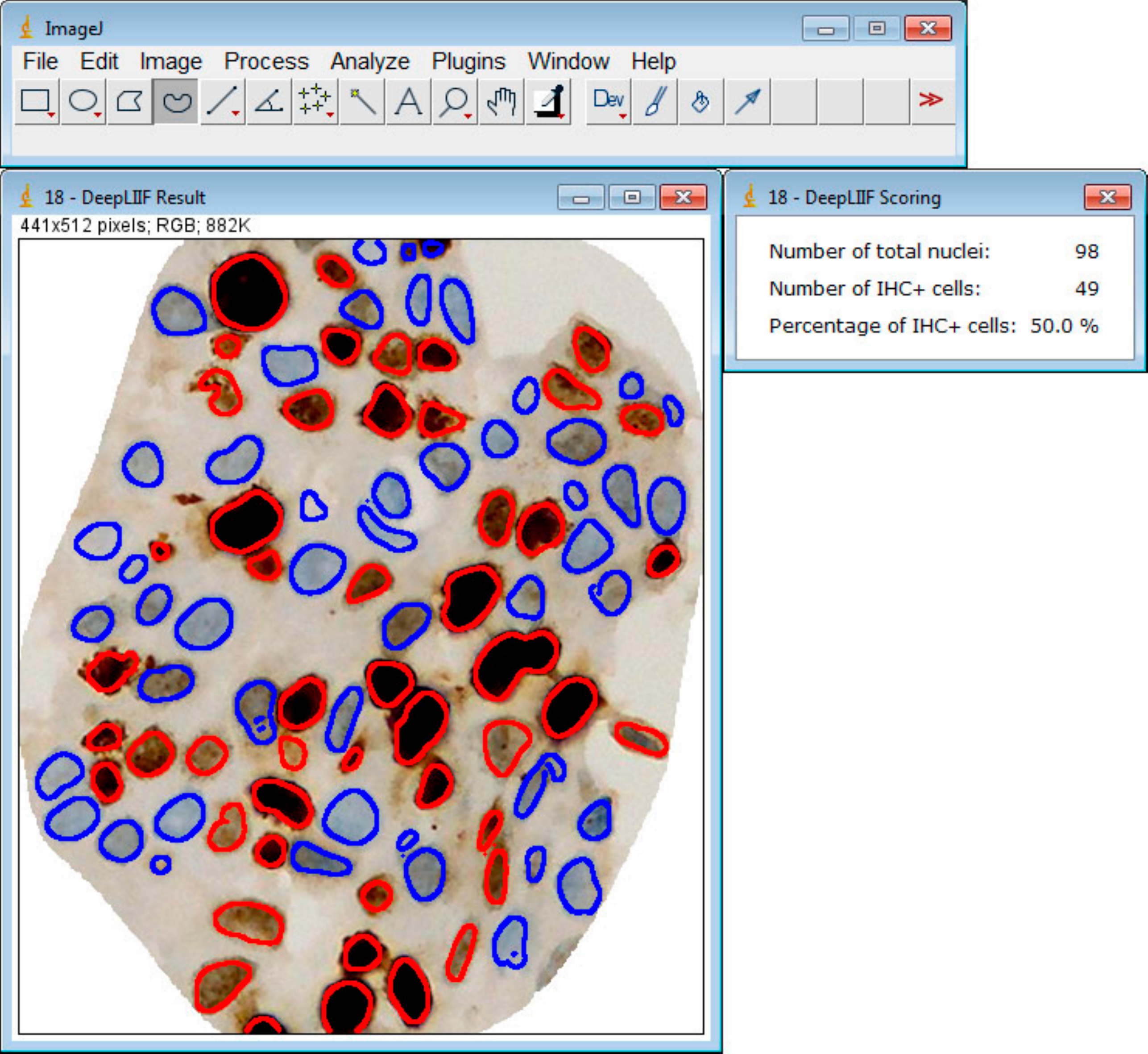}
\caption{DeepLIIF results for ROI.}
\end{subfigure}
\begin{subfigure}{.37333789\textwidth}
\raggedleft
\includegraphics[width=.99\linewidth,keepaspectratio]{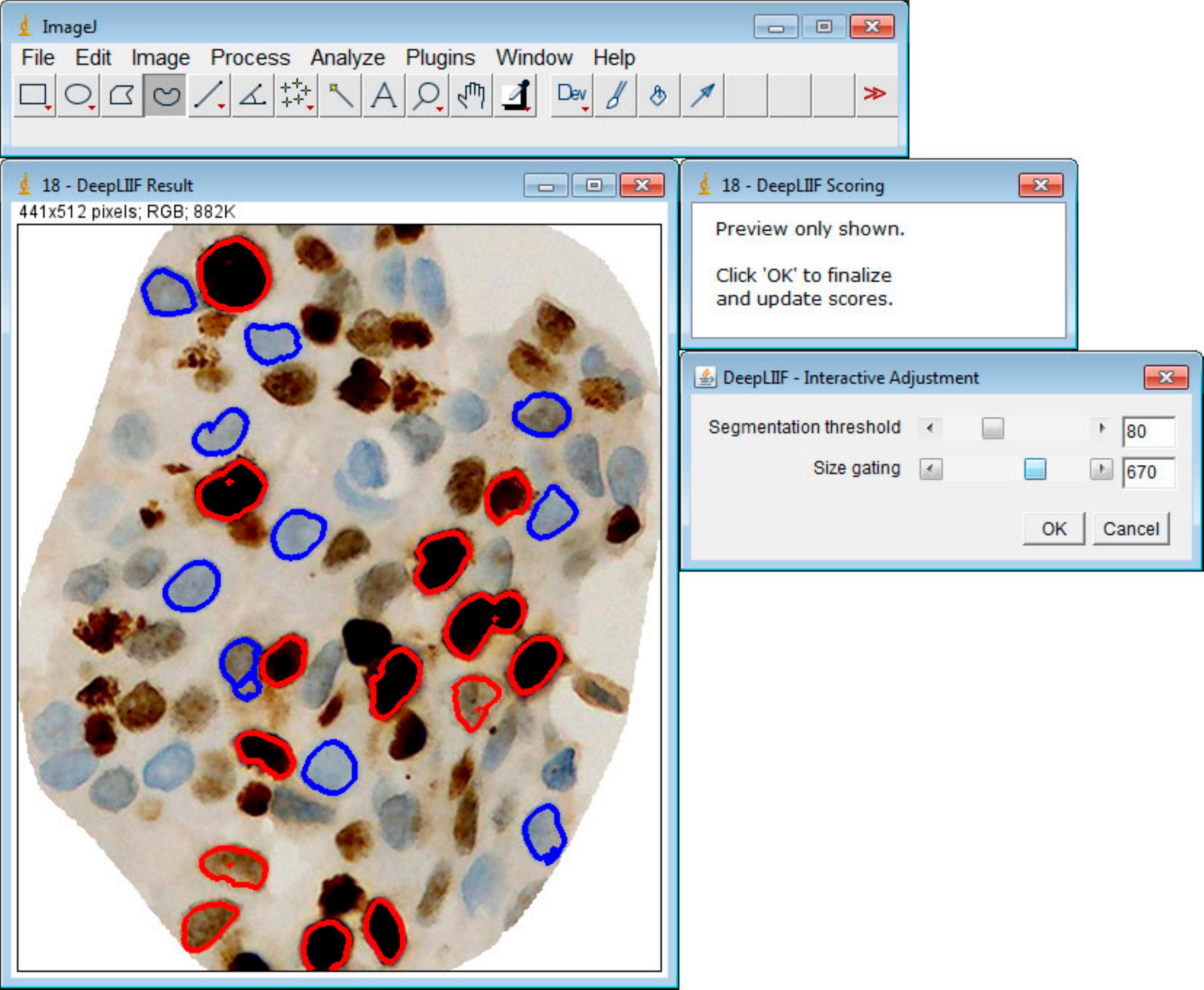}
\caption{Interactive adjustment.}
\end{subfigure}
\caption{ImageJ plugin.  (a) A user has selected a region of interest (ROI), outlined in yellow, using the ImageJ freehand selection tool.  (b) After running the DeepLIIF plugin to submit the image region and obtain the results, the segmentation and scoring for the ROI are displayed.  (c) The user can invoke the interactive adjustment command of the plugin to adjust the results, as is possible on the DeepLIIF website.}
\label{fig:imagej}
\end{figure*}

\subsection*{API Endpoint}
\noindent
The image inference can also be accomplished through an API endpoint, giving users programmatic access to the inferred modalities, segmentation, and quantification results.  This is accessed by posting a multipart-encoded request containing the original image file to the \texttt{/api/infer} endpoint.  The response is JSON encoded data including an array of images with the results of the DeepLIIF model and the quantification scoring values.  The result images are Base64 encoded Portable Network Graphics (PNG) image data, allowing them to be easily viewed or saved locally.  The following query parameters can be passed:
\begin{itemize}
    \item \texttt{resolution} (\textit{string}): resolution used to scan the slide ($10\times$, $20\times$, $40\times$); defaults to $20\times$.
    \item \texttt{pil} (\textit{boolean}): if true, use the Python Imaging Library to load the image instead of Bio-Formats.  For common image formats, this will reduce the time required to read the image on the server.
    \item \texttt{slim} (\textit{boolean}): if true, return only the segmentation result image.  This will reduce the amount of data that is returned.
\end{itemize}

A Python code sample showing how to access and obtain results from this API is given in Code~\ref{code:api}.

\subsection*{ImageJ Plugin}
\noindent
The API endpoint also allows integration of our platform into open source whole slide viewers, such as QuPath/ImageJ, through custom plugins.  We have developed and made available an ImageJ plugin which allows users to easily submit images and obtain results from the DeepLIIF cloud platform.  This plugin provides a user experience that is similar to that of the website, but with the user able to take advantage of the additional tools provided by ImageJ.  For example, Figure~\ref{fig:imagej} shows DeepLIIF being run on a region of interest (ROI) that a user has selected within an image.  All of the inferred image results, along with the scoring results, are saved as local files.

\begin{table*}[t]
\begin{tabular}{|c|}
\hline 
\begin{minipage}{\linewidth}
\small
\begin{lstlisting}[language=Python]
import os
import json
import base64
from io import BytesIO
import requests
from PIL import Image

# Use the sample images from the main DeepLIIF repo
images_dir = './Sample_Large_Tissues'
filename = 'ROI_1.png'

res = requests.post(
    url='https://deepliif.org/api/infer',
    files={
        'img': open(f'{images_dir}/{filename}', 'rb')
    },
    # optional param that can be 10x, 20x (default), or 40x
    params={
        'resolution': '20x'
    }
)

data = res.json()

def b64_to_pil(b):
    return Image.open(BytesIO(base64.b64decode(b.encode())))

for name, img in data['images'].items():
    output_filepath = f'{images_dir}/{os.path.splitext(filename)[0]}_{name}.png'
    with open(output_filepath, 'wb') as f:
        b64_to_pil(img).save(f, format='PNG')

print(json.dumps(data['scoring'], indent=2))

\end{lstlisting}
\end{minipage} \\
\hline
\end{tabular}
\renewcommand{\tablename}{Code}
\caption{Python code sample for using the DeepLIIF cloud API endpoint.}
\label{code:api}
\end{table*}

\section{Limitations and Future Work}
\label{sec:limitations}
\noindent
Currently, we enforce a limit on the size of uploaded images to no larger than $3000\times3000$ pixels. As we continue to improve the performance of our model, we will increase the size of images which are allowed. Moreover, at present we expect users to use our ImageJ/QuPath plugin to open whole slide images, crop regions of interest (normally tumor regions) for IHC scoring, and upload to the DeepLIIF platform. Users can obviously crop regions of interest in other whole slide image viewers and upload these manually as well. Our eventual goal is to incorporate an open source whole slide image viewer (e.g. \mbox{HistomicsUI}, \url{https://digitalslidearchive.github.io/digital_slide_archive/}) into the web interface, which will allow users to upload a whole slide image, annotate one or more regions of interest, and then process those selected regions to obtain the inferred modalities and quantification. 

Our platform currently only supports nuclear markers (such as Ki67, ER, PR) and cytoplasmic markers (e.g. CD3/CD8) that express close to the boundary of the nuclei. In the future, we also plan to support cytoplasmic/membranous IHC protein markers, such as PD-L1, HER2, etc. We invite readers to visit our online platform at \url{https://deepliif.org}, which will be under continuous development as we refine our model and add additional features.

\section*{Acknowledgements}
\noindent
This project was supported by MSK Cancer Center Support Grant/Core Grant (P30 CA008748) and MSK DigITs Hybrid Research Initiative.


\end{document}